%% file: main.tex
\documentclass[10pt,twocolumn,letterpaper]{article}

\usepackage{cvpr}      


\definecolor{cvprblue}{rgb}{0.21,0.49,0.74}
\usepackage[pagebackref,breaklinks,colorlinks,allcolors=cvprblue]{hyperref}

\usepackage{multicol}
\usepackage{multirow}
\usepackage{amssymb}
\usepackage{comment}
\usepackage[accsupp]{axessibility}
\def\paperID{104} 
\def\confName{CVPR}
\def\confYear{2025}

\title{Instruction-augmented Multimodal Alignment for Image-Text and Element Matching}
\renewcommand\footnotemark{}
\author{Xinli Yue$^{*1}$
\hspace{.1in}
JianHui Sun$^{*2}$
\hspace{.1in}
Junda Lu$^{2}$
\hspace{.1in}
Liangchao Yao$^{2}$
\hspace{.1in}
FAN XIA$^{2}$ \\
Tianyi Wang$^{2}$
\hspace{.1in}
Fengyun Rao$^{2}$
\hspace{.1in}
JING LYU$^{2}$
\hspace{.1in}
Yuetang Deng$^{\dagger2}$
\thanks{$^*$Equal contribution.}
\thanks{$^{\dagger}$Corresponding author.} \\
$^{1}$Wuhan University
\hspace{.1in}
$^{2}$WeChat
}
\begin{document}
\maketitle

\begin{abstract}
With the rapid advancement of text-to-image (T2I) generation models, assessing the semantic alignment between generated images and text descriptions has become a significant research challenge. Current methods, including those based on Visual Question Answering (VQA), still struggle with fine-grained assessments and precise quantification of image-text alignment. This paper presents an improved evaluation method named \textbf{I}nstruction-augmented Multimodal Alignment for Image-Text and Element \textbf{Match}ing (iMatch), which evaluates image-text semantic alignment by fine-tuning multimodal large language models. We introduce four innovative augmentation strategies: First, the QAlign strategy creates a precise probabilistic mapping to convert discrete scores from multimodal large language models into continuous matching scores. Second, a validation set augmentation strategy uses pseudo-labels from model predictions to expand training data, boosting the model's generalization performance. Third, an element augmentation strategy integrates element category labels to refine the model's understanding of image-text matching. Fourth, an image augmentation strategy employs techniques like random lighting to increase the model's robustness. Additionally, we propose prompt type augmentation and score perturbation strategies to further enhance the accuracy of element assessments. Our experimental results show that the iMatch method significantly surpasses existing methods, confirming its effectiveness and practical value. Furthermore, our iMatch won first place in the CVPR NTIRE 2025 Text to Image Generation Model Quality Assessment - Track 1 Image-Text Alignment.
\end{abstract}

\begin{figure*}[ht]
	\centering
	\includegraphics[width=1.0\linewidth]{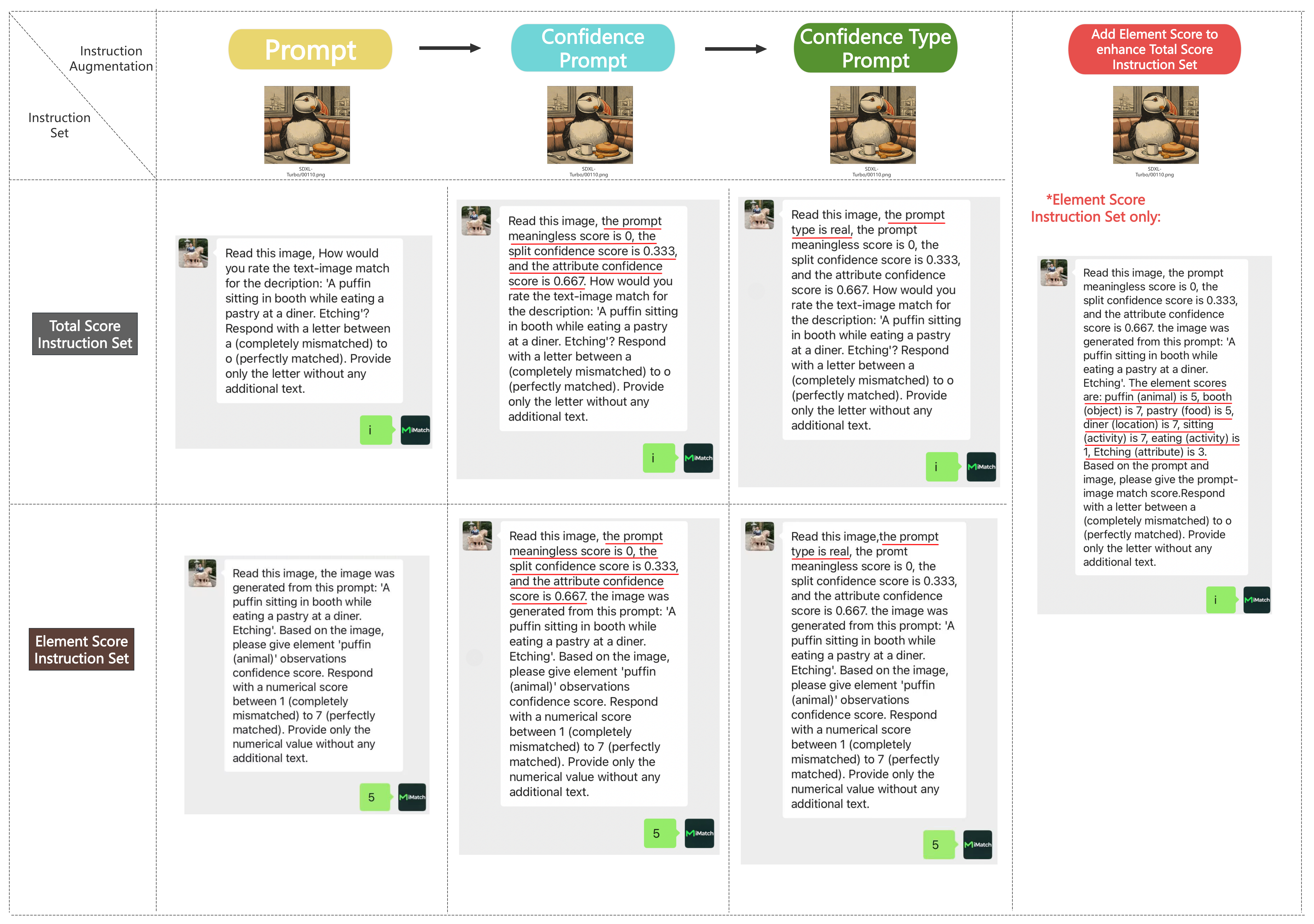}
	\caption{The instruction set augmentation process of the proposed iMatch.}
	\label{fig_prompt_augmentation}
\end{figure*}

\section{Introduction}

In recent years, the rapid development of deep learning technologies has led to significant breakthroughs in text-to-image (T2I) generation models~\cite{dalle,sd1.5,midjourney,dreamina,kandinsky3.0,playgroundv2.5,sd3,hunyuandit,sdxl_turbo,flux}, which have demonstrated powerful image generation capabilities. However, objectively and accurately assessing the semantic alignment of these generated images with text descriptions has increasingly become a critical issue and a significant challenge in the research field.

Metrics based on visual language models~\cite{clip,blip2} assess the semantic matching between image and text by measuring cosine similarity in embedding spaces. Several approaches~\cite{imagereward, pickscore, tifa, gecko} utilize human-annotated data to emulate human judgments of image-text alignment, while others analyze texts broken down into elements evaluated through VQA models. Innovations such as the introduction of question templates improve the alignment of the VQA model with human preferences~\cite{evalmuse}. Although these methods form a foundation for evaluating image-text alignment, they lack a unified approach for comprehensive and detailed matching assessments, missing a systematic exploration of the complex relationships between text and images.

On the other hand, the emergence of Multimodal Large Language Models (MLLMs)~\cite{gpt4,llava,minigpt-4,Qwen-VL,llava1.5,gemini,internvl,llavanext,deepseekvl} in recent years has provided robust technical support for image-text matching tasks. Recently, models such as InternVL2.5~\cite{internvl2.5}, Qwen2.5-VL~\cite{qwen2.5vl}, and the Ovis2~\cite{ovis2} series have shown exceptional performance in multimodal understanding tasks. Nonetheless, further exploration is needed to fully leverage these models for more accurate assessments of image-text matching.

Addressing the challenges highlighted above, this study proposes an enhanced method for comprehensive and fine-grained image-text matching assessment, named \textbf{I}nstruction-augmented Multimodal Alignment for Image-Text and Element \textbf{Match}ing (iMatch), aimed at precisely measuring the alignment between generated images and textual descriptions. Specifically, our approach encompasses several innovative aspects: Firstly, we employ a fine-tuning strategy based on MLLMs, utilizing fine-grained image-text matching annotations from the EvalMuse-40K dataset~\cite{evalmuse} to explicitly guide the model in learning the nuanced correspondences between text and images. To further enhance model performance, we propose four augmentation strategies: (1) QAlign~\cite{qalign} strategy: This strategy defines a mapping function from textual rating levels to specific numerical scores, coupled with a soft mapping of model prediction probabilities, to more accurately convert overall image-text matching scores. (2) Validation set augmentation strategy: We initially use the model to predict on the validation set during training, generating high-quality pseudo-labels, which are then merged back into the original training set for re-training to enhance the model's generalization performance. (3) Element augmentation strategy: During training, we explicitly input element labels as additional features into the user query, helping the model deduce the overall matching score from finer-grained information in a Chain-of-Thought-like manner~\cite{cot}. (4) Image augmentation strategy: We introduce three data augmentation techniques to expand the diversity of training set images and enhance the model's robustness to image variations.

Additionally, in the element matching task, we propose two additional augmentation techniques: (1) Prompt type augmentation: We explicitly integrate the prompt type (real or synthetic) into the user query, aiding the model in distinguishing the intrinsic characteristics of different prompt sources. (2) Score perturbation augmentation: By applying slight random perturbations to the target labels of element matching, we reduce the risk of model overfitting to specific training labels, further improving the model's generalization capabilities. The synergistic effect of these methods has led to outstanding performance on the EvalMuse-40K dataset~\cite{evalmuse} and in the NTIRE 2025 challenge~\cite{ntire2025text}, validating the effectiveness and practicality of our proposed methods in image-text matching tasks.

In summary, our contributions are as follows:
\begin{itemize} 
    \item We present iMatch, an innovative image-text matching method that enhances semantic matching accuracy through fine-tuned multimodal models and strategic augmentations. \item We introduce four augmentation strategies—QAlign, validation set augmentation, element augmentation, and image augmentation—to improve the model's adaptability and generalization in image-text tasks.
    \item We develop two techniques—prompt type augmentation and score perturbation augmentation—to boost model performance and stability. 
    \item Our extensive testing on the EvalMuse-40K dataset and in the NTIRE 2025 challenge shows that iMatch surpasses existing methods across multiple metrics.
\end{itemize}

\section{Related Work}
\subsection{Image-Text Alignment}
Recent advancements in text-to-image (T2I) generation models~\cite{midjourney,dreamina,kandinsky3.0,playgroundv2.5,sd3,hunyuandit,sdxl_turbo,flux,ranni,kolors,mars} have emphasized the importance of accurately assessing the alignment between images and text descriptions. CLIPScore~\cite{clipscore} utilizes a pretrained CLIP~\cite{clip} model to measure cosine similarity between embeddings, providing an initial automatic evaluation. BLIP2Score~\cite{clipscore} follows a similar approach for enhanced evaluation. ImageReward~\cite{imagereward} and PickScore~\cite{pickscore} refine assessments through fine-tuning with extensive human feedback, aligning more closely with human perceptions. TIFA~\cite{tifa} and VQ2~\cite{vq2} break down text into element-level questions answered by VQA models, focusing on detailed matching. FGA-BLIP2~\cite{evalmuse} improves this with tailored question templates to direct VQA models towards crucial text content, enhancing element-level accuracy. Despite these advances, challenges remain in providing a unified framework for detailed and overall image-text alignment. 

\subsection{Multimodal Large Language Models}
In recent years, advanced MLLMs~\cite{gpt4,llava,minigpt-4,Qwen-VL,visualchatgpt,hugginggpt,llava1.5,gemini,internvl,llavanext,deepseekvl} have shown remarkable performance in multimodal tasks. The InternVL2.5 series~\cite{internvl2.5}, based on InternVL 2.0, retains the original architecture but introduces improvements in training, testing, and data quality. The InternVL2.5-MPO model employs Mixed Preference Optimization (MPO) to enhance reasoning in multimodal tasks. The Qwen2.5-VL series~\cite{qwen2.5vl} demonstrates strong visual and linguistic skills, with the Qwen2.5-VL-7B-Instruct model excelling in visual benchmarks and agentive tasks. The Ovis2 models~\cite{ovis2}, successors to Ovis1.6, feature upgraded dataset organization and training, boosting reasoning in smaller models. These developments in MLLMs advance cross-modal understanding and generation, setting the stage for future AI advancements.

\section{Methodology} 
\subsection{Baseline Model} 

We propose a fine-tuning approach based on MLLMs~\cite{internvl2.5,qwen2.5vl,ovis2} to address the fine-grained image-text matching task in image generation quality assessment. Utilizing the EvalMuse-40K dataset~\cite{evalmuse}, which provides extensive fine-grained image-text matching annotations, our method explicitly guides the model in learning the detailed correspondences between text and images. Specifically, given a text description and a corresponding generated image, the model is tasked with predicting both an overall image-text matching score and element-level matching hits.

\subsubsection{Problem Definition} Let the given text description be $P$, and its corresponding generated image be $I$. The EvalMuse-40K dataset~\cite{evalmuse} provides an overall image-text matching score $S_{\text{total}}$ and a set of fine-grained element matching scores $\left\{S_{e_i}\right\}_{i=1}^N$ for each image-text pair $(I, P)$. Here, $S_{\text{total}}$ ranges from [1,5], representing the overall matching degree, while each element matching score $S_{e_i}$ ranges from [0,1], indicating whether a specific element is accurately represented in the image.

Therefore, the task can be formally defined as:
\begin{equation}
    \hat{S}_{\text {total }}=f_{\text {total }}(I, P ; \theta), 
\end{equation}
\begin{equation}
    \quad \hat{y}_{e_i}=1\left[f_{\text {element }}\left(I, P, e_i ; \phi\right)>\tau\right], \quad i=1,2, \ldots, N,
\end{equation}
where $f_{\text{total}}$ is the model predicting the overall image-text matching score, $f_{\text{element}}$ is the model predicting element matching scores, $\theta$ and $\phi$ are model parameters, and $\tau$ is the threshold for determining element hits.

\begin{figure}[t]
	\centering
	\includegraphics[width=1.0\linewidth]{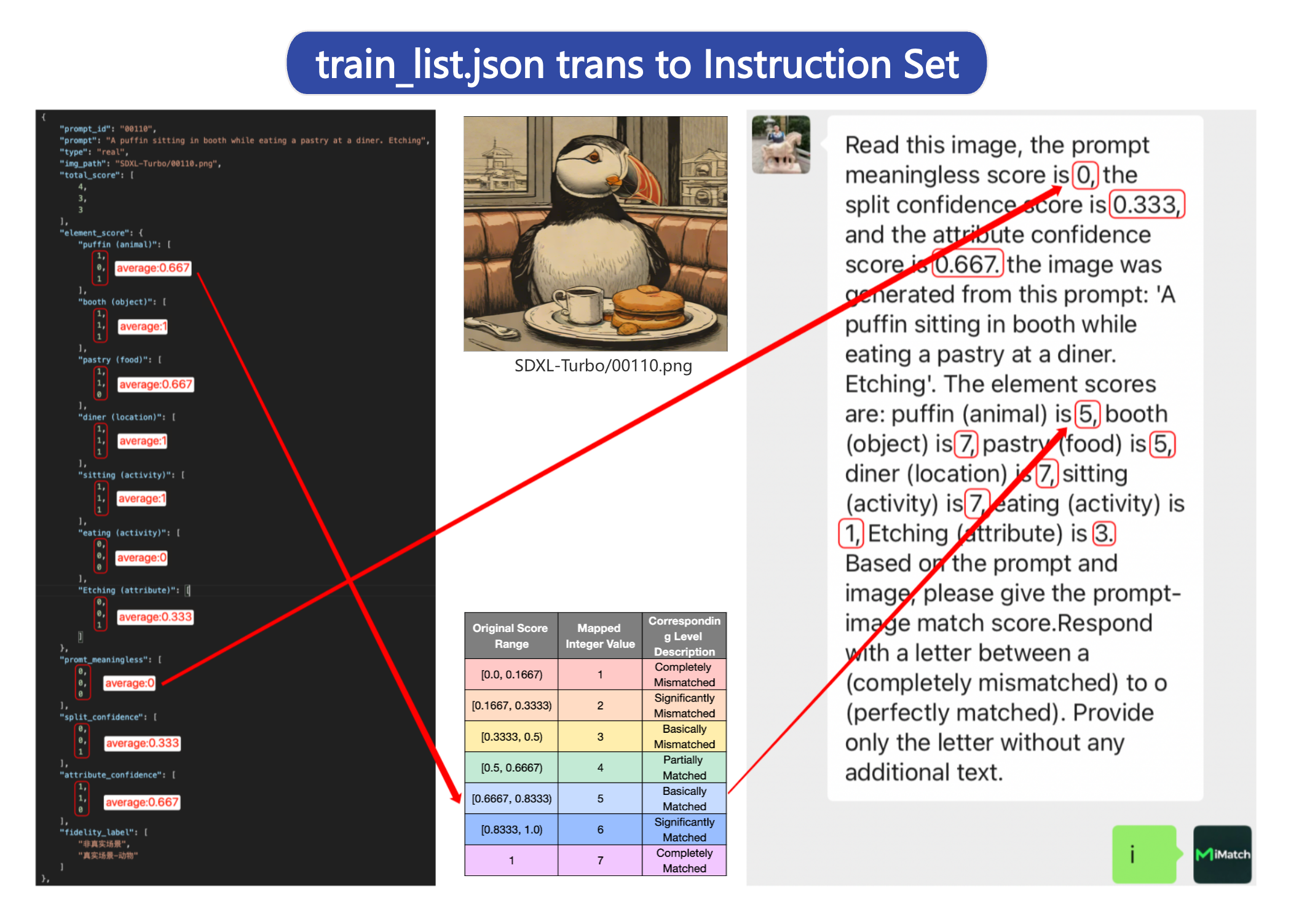}
	\caption{The construction process of the element instruction set augmentation. We map elements to integers between 1 and 7 and incorporate them into the user query. Additionally, confidence scores are also integrated into the user query.}
	\label{fig_instruction_set}
\end{figure}

\subsubsection{Instructional Fine-tuning Strategy}
To enhance the model's capability for learning fine-grained image-text correspondences, we have designed a specific instructional fine-tuning strategy that transforms the task of predicting image-text matching scores into a classification problem. The steps are as follows:

For the overall image-text matching score $S_{\text{total}}$, we first perform a linear scaling:
\begin{equation}
    S_{\text{total}}' = \text{round}\left(\frac{S_{\text{total}} - 1}{4} \times 14 + 1\right).
\end{equation}    

At this point, $S_{\text{total}}'$ is transformed into a discrete set of integers \{1, 2, \ldots, 15\}. Subsequently, we map this integer range to the alphabet set \{a, b, \ldots, o\}, which serves as the target labels for multimodal model instructional fine-tuning. During the inference phase, the model’s predicted letter labels are mapped back to the 1--15 range and linearly scaled to the original 1--5 range as the final prediction result.

For the element matching task, as illustrated in Figure \ref{fig_instruction_set}, we discretize the element matching scores $S_{e_i}$ into 7 categories:
\begin{equation}
    S_{e_i}' = \text{round}(S_{e_i} \times 6) + 1.
\end{equation}
    
During inference, we use a threshold $\tau = 3$, converting the predicted categories into a binary classification task, where scores greater than 3 are considered hits (1), and others are considered misses (0).

Additionally, we incorporate other information provided by the EvalMuse-40K dataset~\cite{evalmuse} such as prompt meaninglessness, split confidence, and attribute confidence into the problem formulation, as shown in Figure 
\ref{fig_instruction_set}. Through this instruction-based fine-tuning strategy, we significantly enhance the multimodal model's performance on image-text matching tasks, especially in terms of element-level recognition and judgment.

\subsection{Image-Text Matching Augmented Model}
\label{image-text matching augmented model}

\begin{figure}[t]
	\centering
	\includegraphics[width=1.0\linewidth]{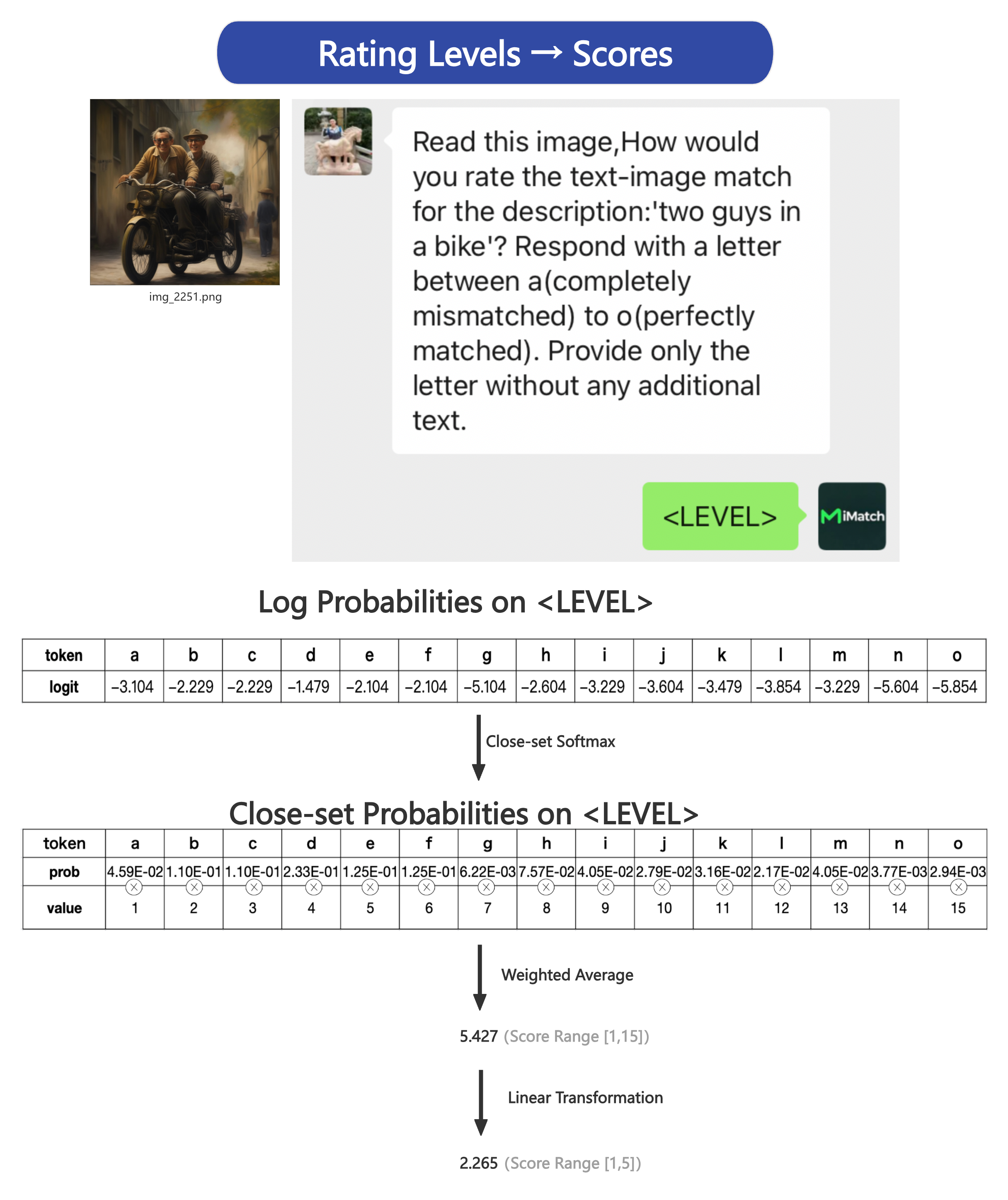}
	\caption{The inference process of the image-text matching augmented model. During inference, we extract closed-set probabilities for rating levels and perform a weighted average to obtain the MLLM-predicted score.}
	\label{fig_qalign}
\end{figure}

\subsubsection{QAlign Augmentation}
To further enhance the accuracy of image-text matching tasks, we introduce a probability distribution-based augmentation strategy to meticulously simulate human scoring behavior, as illustrated in Figure \ref{fig_qalign}. Specifically, we adopt a post-processing strategy similar to the QAlign method~\cite{qalign}, which converts the model output from scoring levels to more precise final matching scores.

Firstly, we define a mapping function $G$ from letters to numeric scores:
\begin{equation}
    G: l_i \rightarrow i, i \in \{1, 2, \ldots, 15\},
\end{equation}
where $\{l_i|_{1}^{15}\} = \{a,b,...,o\}$, for example, scoring level \( a \) corresponds to numeric score 1, while level \( o \) corresponds to numeric score 15.

Next, we calculate the probabilities \( p_{l_i} \) predicted by the MLLMs. Specifically, if the logits output by the language model for each scoring level \( l_i \) are \( x_{l_i} \), we use the closed-set softmax function to compute the probability distribution for each level:
\begin{equation}
    p_{l_i} = \frac{e^{x_{l_i}}}{\sum_{j=1}^{15} e^{x_{l_j}}},
\end{equation}
where \( \sum_{i=1}^{15} p_{l_i} = 1 \).

Consequently, the model’s final continuous predicted score $\hat{S}_{\text {total }}$ is:
\begin{equation}
    \hat{S}_{\text {total }} = \sum_{i=1}^{15} p_{l_i} \cdot G(l_i).
\end{equation}

\subsubsection{Validation Set Augmentation}

The NTIRE 2025 competition has a development phase with a validation set, and a final phase with a test set. To further enhance the model's generalization ability during the final test phase, we design a pseudo-labeling strategy that fully utilizes the validation set data to augment the training process.  In the development phase, we first use a trained model to predict the validation set, generating pseudo labels, denoted as \( \hat{y}_v \). Then, in the final phase, we combine the pseudo-labeled validation set with the original training set to construct an augmented training dataset for subsequent model retraining.

The new training objective can be expressed as:
\begin{equation}
    \mathcal{L}_{\text{enhanced}}(\theta) = \mathcal{L}_{\text{train}}(y_t, \hat{y}_t; \theta) + \mathcal{L}_{\text{pseudo}}(\hat{y}_v, \hat{y}_v'; \theta),
\end{equation}

where \( \mathcal{L}_{\text{train}} \) denotes the loss on the original training set, \( \mathcal{L}_{\text{pseudo}} \) denotes the loss supervised by pseudo labels, and \( \hat{y}_v' \) is the model's prediction on the validation set samples.

\begin{figure}[t]
	\centering
	\includegraphics[width=1.0\linewidth]{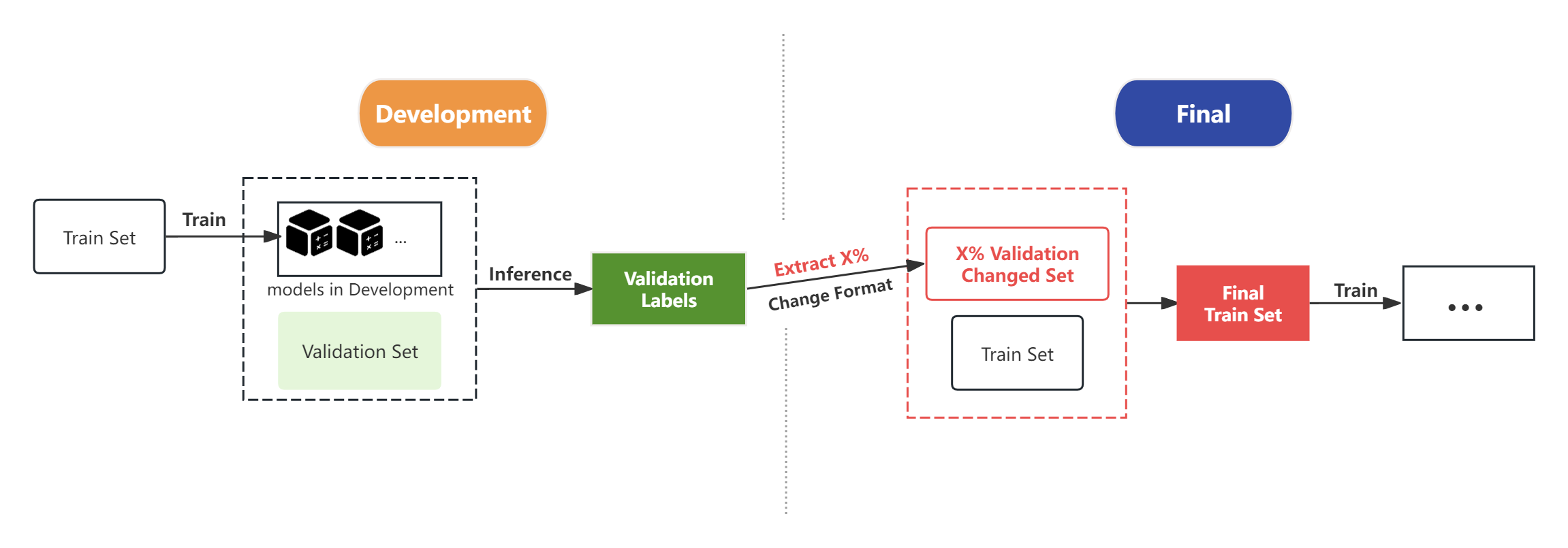}
	\caption{The process of validation set augmentation. We use the model trained on the training set to generate pseudo-labels for the validation set during the development phase, and then use these pseudo-labeled validation data to augment the training dataset.}
	\label{fig_validation_set_augmentation}
\end{figure}

\subsubsection{Element Augmentation}
To further enhance the model's understanding and judgment capabilities for image-text matching, we propose a strategy based on element feature augmentation. Specifically, as shown in Figure \ref{fig_instruction_set} during the training phase, element-level scores are explicitly embedded into the user query as additional features to facilitate a reasoning process similar to the Chain-of-Thought (CoT)~\cite{cot}.

However, during the testing phase, due to the lack of real element labels, we cannot directly employ the same augmentation method. Therefore, we adopt a pseudo-label prediction strategy. Initially, we use the model trained in the Section \ref{element matching augmented model} to predict pseudo-labels for the element level \( \hat{y}_{e_i} \) in the test set. Specifically, the model first predicts element scores:
\begin{equation}
    \hat{S}_{e_i} = f_{\text{element}}(I, P, e_i; \phi).
\end{equation}

Subsequently, these predicted element scores \( \hat{S}_{e_i} \) are embedded into the input of the image-text matching task, forming a prompt with pseudo-labeled element scores to predict the final image-text matching score $\hat{S}_{\text {total }}$:
\begin{equation}
    \hat{S}_{\text {total }} = f_{\text{total}}(I, P, \hat{S}_{e_i}; \theta).
\end{equation}

\subsubsection{Image Augmentation}
\begin{figure}[t]
	\centering
	\includegraphics[width=1.0\linewidth]{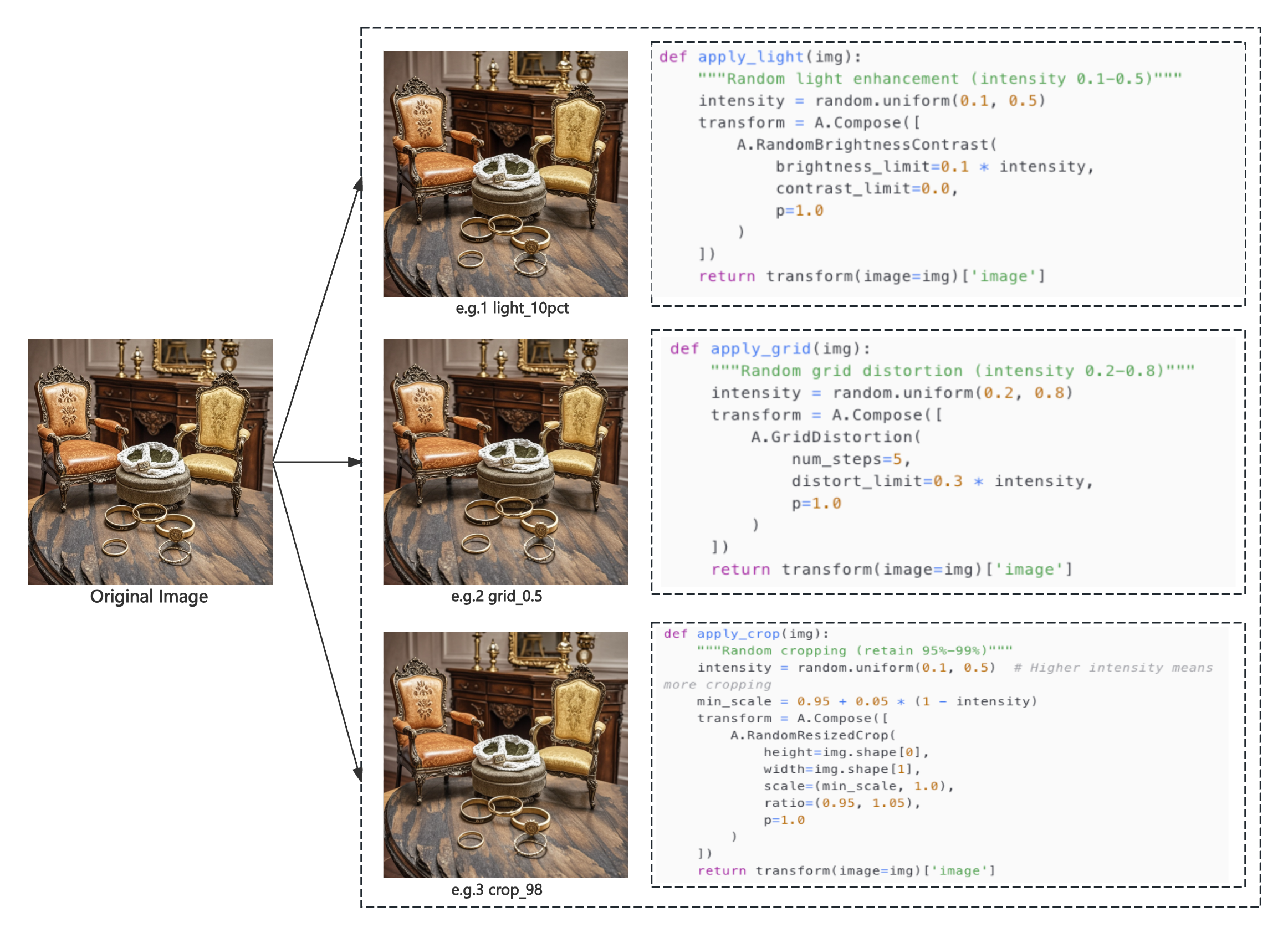}
	\caption{Examples of image augmentation. We employ three types of augmentations: Random Lighting Augmentation, Random Grid Distortion, and Random Crop Augmentation.}
	\label{fig_image_augmentation}
\end{figure}

To further enhance the model's generalization performance and improve robustness to image variations, we introduce an image data augmentation strategy, as illustrated in Figure \ref{fig_image_augmentation}. Specifically, we randomly select 10\% of the samples from the training set and apply one of three different types of data augmentation methods. These augmented samples are then used in conjunction with the original data for training, effectively enriching the diversity of the training dataset. The specific data augmentation methods include:
\paragraph{Random Lighting Augmentation.} We randomly adjust the brightness of images to simulate variations in lighting conditions, thereby improving the model's robustness to changes in illumination. Defined as:
\begin{equation}
    I_{\text{light}} = T_{\text{brightness}}(I, \alpha), \alpha \sim U(0.1, 0.5),
\end{equation}
where \( I \) represents the original image, and \( \alpha \) is a randomly sampled lighting intensity factor.
\paragraph{Random Grid Distortion.} We apply random grid deformations to the image to enhance the model's robustness to spatial transformations. Specifically defined as:
\begin{equation}
    I_{\text{grid}} = T_{\text{grid}}(I, \beta), \beta \sim U(0.2, 0.8),
\end{equation}
where \( \beta \) is a random factor controlling the degree of grid distortion.
\paragraph{Random Crop Augmentation.} We randomly crop images to simulate partial occlusions or changes in perspective. Specifically defined as:
\begin{equation}
    I_{\text{crop}} = T_{\text{crop}}(I, \gamma), \gamma \sim U(0.1, 0.5),
\end{equation}
where \( \gamma \) determines the cropping scale.

The above augmentation strategies are implemented as follows: for a randomly selected 10\% subset \( D_{\text{subset}} \) from the training dataset \( D_{\text{train}} \):
\begin{equation}
\begin{split}
D_{\text{augmented}} 
&= \{ T(I) \,\mid\, I \in D_{\text{subset}}, \\
 &\qquad\quad\, T \in \{T_{\text{brightness}}, T_{\text{grid}}, T_{\text{crop}}\} \}.
\end{split}
\end{equation}

Subsequently, we merge the enhanced dataset with the original dataset to construct the final augmented training set:
\begin{equation}
    D_{\text{final}} = D_{\text{train}} \cup D_{\text{augmented}}.
\end{equation}

\subsection{Element Matching Augmented Model}
\label{element matching augmented model}
Building upon the baseline model, we further propose an element matching augmented model aimed at improving the model's fine-grained understanding and predictive accuracy regarding image-text element correspondences. Specifically, we introduce two augmentation strategies, including the incorporation of prompt type information and the introduction of score perturbation.

\subsubsection{Prompt Type Augmentation}
To more richly characterize the problem features in the element matching task, we propose integrating prompt type information from the EvalMuse-40K dataset—either "real" or "synthetic" (generated using GPT-4)—explicitly into the user query. By explicitly introducing prompt type information \( t \in \{ \text{real}, \text{synth} \} \), we expand the input formulation, resulting in an enhanced element matching task representation:
\begin{equation}
    \hat{y}_{e_i} = f_{\text{element}}(I, P, e_i, t; \phi').
\end{equation}

\subsubsection{Score Perturbation}
To further enhance the generalization capability of the element matching model and prevent overfitting to specific score annotations in the training data, we propose a perturbation strategy based on element labels. Specifically, during the training phase, we apply slight perturbations to the mapped element labels, which have a discrete numerical range of \{1, 2, 3, 4, 5, 6, 7\}, to increase the diversity of the training data.

Let the discretized score of element \( e_i \) in the training set be denoted as \( S_{e_i}^{\text{(discrete)}} \). We introduce a perturbation factor \( \epsilon \), and add a random perturbation \( \delta \in \{-\epsilon, \epsilon\} \) to each element's discretized score. The perturbed element score \( S_{e_i}^{\text{(perturbed)}} \) can then be expressed as:
\begin{equation}
    S_{e_i}^{\text{(perturbed)}} = S_{e_i}^{\text{(discrete)}} + \delta, \quad \delta \sim U\{-\epsilon, \epsilon\}.
\end{equation}

\begin{figure}[t]
	\centering
	\includegraphics[width=1.0\linewidth]{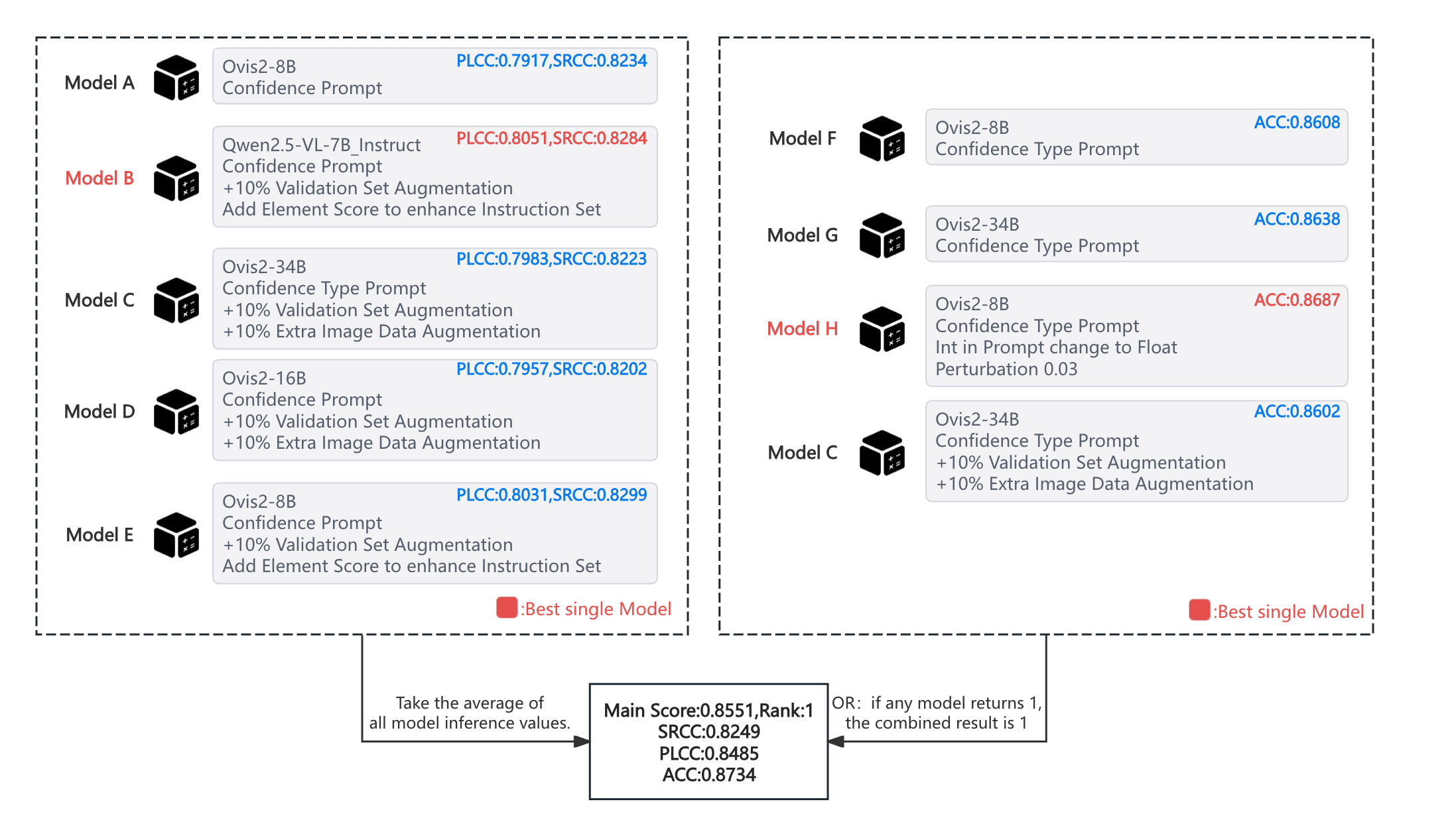}
	\caption{Model Ensemble. We ensemble five image-text matching augmented models and four element matching augmented models to improve the model's generalization performance.}
	\label{fig_overview}
\end{figure}

\begin{table}[t]
\begin{center}
\caption{Performance comparison of different methods on the EvalMuse-40K validation set for image-text matching tasks.}
\label{tab_validation_ps}
\begin{tabular}{@{}ccc@{}}
\toprule[1.0pt]
Method                 & SRCC            & PLCC            \\ \midrule
CLIPScore~\cite{clipscore}     & 0.2993          & 0.2933          \\
BLIPv2Score~\cite{clipscore} & 0.3583          & 0.3348          \\
ImageReward~\cite{imagereward} & 0.4655          & 0.4585          \\
PickScore~\cite{pickscore}     & 0.4399          & 0.4328          \\
HPSv2~\cite{hpsv2}         & 0.3745          & 0.3657          \\
VQAScore~\cite{vqascore}      & 0.4877          & 0.4841          \\
FGA-BLIP2~\cite{evalmuse}         & 0.7742          & 0.7722          \\
InternVL2.5-8B-MPO (zero-shot) &0.7262 &0.6742 \\
iMatch (ours)           & \textbf{0.8304} & \textbf{0.8294} \\ \bottomrule[1.0pt]
\end{tabular}
\end{center}
\end{table}

\section{Experiments}

\subsection{Experimental Setup}

\textbf{Dataset.} Our experimental studies employ the EvalMuse-40K dataset~\cite{evalmuse}, which consists of images generated by over 20 different text-to-image (T2I) generation models based on approximately 4,000 prompts. Each image-text pair is annotated with two levels of scoring information: an overall prompt-level image-text matching score and a fine-grained element-level matching score. The dataset is divided into a training set, a validation set, and a test set. The training set contains about 30,000 image-text pairs for model parameter learning; the validation set includes about 10,000 pairs for tuning model parameters and selecting hyperparameters; the test set comprises approximately 5,000 pairs for final performance evaluation.

\textbf{Models.} We fine-tune a variety of advanced MLLMs, including InternVL2.5-8B-MPO~\cite{internvl2.5}, Qwen2.5-VL-7B-Instruct~\cite{qwen2.5vl}, Ovis2-8B~\cite{ovis2}, Ovis2-16B~\cite{ovis2}, and Ovis2-34B~\cite{ovis2}, to comprehensively validate the effectiveness and generalization capability of our proposed methods.

\textbf{Training Parameters.} We implement our experiments using the ms-swift framework~\cite{ms-swift}, employing Low-Rank Adaptation (LoRA)~\cite{lora} for fine-tuning, where the LoRA rank is set to 16. Additionally, we unfreeze the Vision Transformer (ViT)~\cite{vit} and the cross-modal aligner to fully leverage cross-modal information. Specific training parameters include a maximum image pixel count of 200,704, an initial learning rate of 4e-5, a weight decay factor of 0.01, a learning rate warm-up proportion of 0.03, and a learning rate decay strategy using cosine annealing. All models are trained for one epoch to ensure fair and consistent experimental conditions.

\textbf{Evaluation Metrics.} To comprehensively assess model performance, we utilize three metrics: the Spearman Rank Correlation Coefficient (SRCC), Pearson Linear Correlation Coefficient (PLCC), and element-level accuracy (ACC). These are combined into a final performance score, calculated as Final Score = PLCC/4 + SRCC/4 + ACC/2.

\textbf{Comparative Approaches.} The comparative schemes include CLIPScore~\cite{clipscore}, BLIPv2Score~\cite{clipscore}, ImageReward~\cite{imagereward}, PickScore~\cite{pickscore}, HPSv2~\cite{hpsv2}, VQAScore~\cite{vqascore}, TIFA~\cite{tifa}, VQ2~\cite{vq2}, PN-VQA~\cite{evalmuse}, and FGA-BLIP2~\cite{evalmuse}.

\begin{table}[t]
\begin{center}
\caption{Performance comparison of different methods on the EvalMuse-40K validation set for element matching tasks. * indicates that the method uses a fixed step search (0.01) to select the optimal binary classification threshold, aiming to maximize overall accuracy.}
\label{tab_validation_acc}
\setlength{\tabcolsep}{1.4mm}{
\begin{tabular}{@{}cccc@{}}
\toprule[1.0pt]
Method                           & MLLMs                  & ACC             \\ \midrule
\multirow{3}{*}{TIFA~\cite{tifa}} & LLaVA1.6~\cite{llavanext}      & 0.6210          \\
                                 & mPLUG-Owl3~\cite{mplugowl3}    & 0.6450          \\
                                 & Qwen2-VL~\cite{qwen2.5vl}      & 0.6450          \\ \midrule
\multirow{3}{*}{VQ2*{[}45{]}}    & LLaVA1.6~\cite{llavanext}      & 0.6750          \\
                                 & mPLUG-Owl3~\cite{mplugowl3}    & 0.6640          \\
                                 & Qwen2-VL~\cite{qwen2.5vl}      & 0.6790          \\ \midrule
\multirow{3}{*}{PN-VQA*~\cite{evalmuse}}         & LLaVA1.6~\cite{llavanext}      & 0.6610          \\
                                 & mPLUG-Owl3~\cite{mplugowl3}    & 0.6760          \\
                                 & Qwen2-VL~\cite{qwen2.5vl}      & 0.6820          \\ \midrule
FGA-BLIP2~\cite{evalmuse}                        & BLIP2~\cite{blip2}         & 0.7680          \\ \midrule 
zero-shot & InternVL2.5-8B-MPO~\cite{internvl2.5} & 0.7681          \\ \midrule
\multirow{3}{*}{iMatch (ours)}    & InternVL2.5-8B-MPO~\cite{internvl2.5}     & 0.8284          \\
                                 & Qwen2.5-VL-7B-Instruct~\cite{qwen2.5vl} & 0.7948          \\
                                 & Ovis2-8B~\cite{ovis2}               & \textbf{0.8317} \\ \bottomrule[1.0pt]
\end{tabular}
}
\end{center}
\end{table}

\subsection{Experimental Results}
\paragraph{Results on Validation Set.} 

We conduct comparative experiments on the EvalMuse-40K validation set against several mainstream methods to evaluate the effectiveness of our proposed iMatch. The performance results are presented in Tables~\ref{tab_validation_ps} and~\ref{tab_validation_acc}.

As shown in Table~\ref{tab_validation_ps}, iMatch achieves significant gains in both SRCC and PLCC, outperforming all baselines. Compared to traditional metrics like CLIPScore, it improves SRCC and PLCC by over 0.533 and 0.539, respectively. Even against the strongest fine-tuned baseline, FGA-BLIP2, iMatch achieves absolute improvements of 0.056 (SRCC) and 0.057 (PLCC), demonstrating the effectiveness of our targeted fine-tuning strategy in capturing fine-grained semantic alignment.

We also report zero-shot results using the same backbone (InternVL2.5-8B-MPO). While the zero-shot model performs well, our fine-tuned iMatch surpasses it by a notable margin, highlighting the importance of task-aware fine-tuning.

In Table~\ref{tab_validation_acc}, iMatch also achieves the best accuracy on the element matching task, outperforming both baseline and zero-shot models. This further validates the effectiveness of our approach in improving both global and detailed image-text alignment.

\begin{table}[t]
\begin{center}
\caption{Results of the NTIRE 2025 Text to Image Generation Model Quality Assessment - Track 1 Image-Text Alignment.}
\label{tab_rank}
\setlength{\tabcolsep}{0.9mm}{
\begin{tabular}{@{}cccccc@{}}
\toprule[1.0pt]
Rank & Team       & Main Score      & SRCC            & PLCC            & ACC             \\ \midrule
1    & IH-VQA (ours)     & \textbf{0.8551} & \textbf{0.8249} & \textbf{0.8485} & \textbf{0.8734} \\
2    & Evalthon   & 0.8426          & 0.8002          & 0.8321          & 0.8691          \\
3    & HCMUS      & 0.8381          & 0.8101          & 0.8306          & 0.8559          \\
4    & MICV       & 0.8221          & 0.7864          & 0.8050          & 0.8485          \\
5    & SJTU-MMLab & 0.8158          & 0.7729          & 0.8029          & 0.8438          \\
6    & SJTUMM     & 0.8062          & 0.7563          & 0.7993          & 0.8346          \\
7    & WT         & 0.7913          & 0.7413          & 0.7740          & 0.8249          \\
8    & YAG        & 0.7777          & 0.7143          & 0.7456          & 0.8255          \\
9    & SPRank     & 0.7604          & 0.6899          & 0.7280          & 0.8119          \\
10   & AIIG       & 0.7386          & 0.6574          & 0.7073          & 0.7949          \\
11   & Joe1007    & 0.7359          & 0.6572          & 0.7041          & 0.7912          \\
12   & iCOST      & 0.7350          & 0.6630          & 0.7040          & 0.7865          \\ \bottomrule[1.0pt]
\end{tabular}
}
\end{center}
\end{table}

\begin{table*}[ht]
\begin{center}
\caption{Results of the ablation study for each component of the image-text matching augmented model on the EvalMuse-40K test set.}
\label{tab_ablation_ps}
{
\begin{tabular}{@{}ccccccccccc@{}}
\toprule[1.0pt]
\multirow{2}{*}{Base} & \multirow{2}{*}{\begin{tabular}[c]{@{}c@{}}QAlign\\      Aug.\end{tabular}} & \multirow{2}{*}{\begin{tabular}[c]{@{}c@{}}Validation\\      Aug.\end{tabular}} & \multirow{2}{*}{\begin{tabular}[c]{@{}c@{}}Element\\      Aug.\end{tabular}} & \multirow{2}{*}{\begin{tabular}[c]{@{}c@{}}Image\\      Aug.\end{tabular}} & \multicolumn{2}{c}{InternVL2.5-8B-MPO} & \multicolumn{2}{c}{Qwen2.5-VL-7B-Instruct} & \multicolumn{2}{c}{Ovis2-8B}      \\ \cmidrule(l){6-7} \cmidrule(l){8-9} \cmidrule(l){10-11}
                      &                                                                             &                                                                                 &                                                                              &                                                                            & SRCC               & PLCC              & SRCC                 & PLCC                & SRCC            & PLCC            \\ \cmidrule(r){1-11}
\checkmark                     & $\times$                                                                           & $\times$                                                                               & $\times$                                                                            & $\times$                                                                          & 0.7626             & 0.7523            & 0.7696               & 0.7381              & 0.7920          & 0.7757          \\
\checkmark                     & \checkmark                                                                           & $\times$                                                                               & $\times$                                                                            & $\times$                                                                          & 0.8049             & 0.7731            & 0.7922               & 0.7667              & 0.8047          & 0.7939          \\
\checkmark                     & \checkmark                                                                           & \checkmark                                                                               & $\times$                                                                            & $\times$                                                                          & 0.8121             & 0.7792            & 0.7957               & 0.7656              & 0.8234          & 0.7917          \\
\checkmark                     & \checkmark                                                                           & \checkmark                                                                               & \checkmark                                                                            & $\times$                                                                          & 0.8213             & \textbf{0.7929}   & \textbf{0.8284}      & \textbf{0.8051}     & \textbf{0.8299} & \textbf{0.8031} \\
\checkmark                     & \checkmark                                                                           & \checkmark                                                                               & $\times$                                                                            & \checkmark                                                                          & \textbf{0.8285}    & 0.7846            & 0.8014               & 0.7716              & 0.8124          & 0.7797          \\
\checkmark                     & $\times$                                                                           & $\times$                                                                               & $\times$                                                                            & \checkmark                                                                          & 0.7697             & 0.7646            & 0.7685               & 0.7428              & 0.7967          & 0.7821          \\ \bottomrule[1.0pt]
\end{tabular}
}
\end{center}
\end{table*}

\begin{table*}[ht]
\begin{center}
\caption{Results of the ablation study for each component of the element matching augmented model on the EvalMuse-40K test set.}
\label{tab_ablation_acc}
\setlength{\tabcolsep}{1.8mm}{
\begin{tabular}{@{}cccccc@{}}
\toprule[1.0pt]
\multirow{2}{*}{Base} & \multirow{2}{*}{\begin{tabular}[c]{@{}c@{}}Type\\      Aug.\end{tabular}} & \multirow{2}{*}{\begin{tabular}[c]{@{}c@{}}Score\\      Perturbation\end{tabular}} & \multicolumn{1}{l}{InternVL2.5-8B-MPO} & \multicolumn{1}{l}{Qwen2.5-VL-7B-Instruct} & \multicolumn{1}{l}{Ovis2-8B} \\ \cmidrule(l){4-6} 
                      &                                                                           &                                                                                    & ACC                                    & ACC                                        & ACC                          \\ \cmidrule(r){1-6}
\checkmark                     & $\times$                                                                         & $\times$                                                                                  & 0.8224                                 & 0.8279                                     & 0.8550                       \\
\checkmark                     & \checkmark                                                                         & $\times$                                                                                  & 0.8393                                 & 0.8342                                     & 0.8608                       \\
\checkmark                     & $\times$                                                                         & \checkmark                                                                                  & 0.8500                                 & 0.8355                                     & 0.8644                       \\
\checkmark                     & \checkmark                                                                         & \checkmark                                                                                  & \textbf{0.8505}                        & \textbf{0.8367}                            & \textbf{0.8687}              \\ \bottomrule[1.0pt]
\end{tabular}
}
\end{center}
\vspace{-0.05 in}
\end{table*}

\begin{table}[t]
\begin{center}
\caption{Inference latency and throughput of iMatch on RTX4090D with images up to 1024$\times$1024 resolution. Batch size is fixed at 4, as larger batches result in OOM.}
\label{tab_cost}
\setlength{\tabcolsep}{0.2mm}{\begin{tabular}{@{}ccc@{}}
\toprule[1.0pt]
Model                  & Latency (s) & Throughput   (image/s) \\ \midrule
InternVL2.5-8B-MPO     & 0.51        & 7.78                     \\
Qwen2.5-VL-7B-Instruct & 0.99        & 4.03                     \\
Ovis2-8B               & 0.54        & 7.46                     \\ \bottomrule[1.0pt]
\end{tabular}}
\end{center}
\end{table}

\paragraph{Results on NTIRE 2025 Challenge.} 

We validated the effectiveness of our method through the NTIRE 2025 Challenge on Text-to-Image Generation Model Quality Assessment – Track 1 Image-Text Alignment. As shown in Figure~\ref{fig_overview}, our final solution leveraged a robust model ensemble strategy that played a key role in its success.

Table~\ref{tab_rank} summarizes the results. Our iMatch method ranked first, outperforming all competitors across all key metrics (SRCC, PLCC, ACC). Compared to the second-place Evalthon team, iMatch improved the Main Score by 0.0125, and achieved gains of 0.0247, 0.0164, and 0.0043 in SRCC, PLCC, and ACC, respectively. The performance margin was even larger relative to other teams, highlighting the strength and consistency of our approach.

\subsection{Ablation Study}
\paragraph{Results on Image-Text Matching Augmented Model.} 

To evaluate the effectiveness of the four augmentation strategies proposed in Section~\ref{image-text matching augmented model}, we conduct ablation experiments on the EvalMuse-40K test set. The results are shown in Table~\ref{tab_ablation_ps}. The introduction of the QAlign strategy significantly improved accuracy in image-text matching. The addition of the validation set augmentation strategy further enhanced the model's generalization performance, as indicated by improvements in both SRCC and PLCC metrics. The element augmentation strategy, which incorporates element-level information, notably increased the model's understanding of fine-grained aspects of image-text relations, achieving the highest improvements. Conversely, employing the image augmentation strategy alone showed effectiveness in enhancing the model’s robustness to image variations, although the element augmentation strategy proved more effective for detailed information capture.

\paragraph{Results on Element Matching Augmented Model.} To assess the impact of the two augmentation strategies proposed in Section~\ref{element matching augmented model}, we conduct ablation experiments on the EvalMuse-40K test set. The results are presented in Table~\ref{tab_ablation_acc}. Introducing the prompt type augmentation strategy significantly improved this accuracy, highlighting its effectiveness in aiding the model's understanding of image-text element relationships, thereby boosting predictive performance. Further incorporating the score perturbation strategy yielded the best performance, demonstrating its value in enhancing model generalization and robustness by introducing moderate label variations.

\subsection{Inference Cost}
In Table \ref{tab_cost}, we report the inference latency and throughput of iMatch on 4 RTX 4090D GPUs. With a fixed batch size of 4 and input resolution up to 1024×1024, the fastest model achieves a latency of 0.51 seconds and a throughput of 7.78 images per second, demonstrating efficient large-scale image processing and potential for real-world deployment.

\section{Conclusion}
In this paper, we tackle the challenges of image quality and semantic matching in text-to-image generation models with a multimodal approach called iMatch, which quantifies both overall and detailed matching relationships between images and text descriptions. We developed a framework using instructional fine-tuning, alongside several innovative strategies such as QAlign, validation set augmentation, element augmentation, and image augmentation. Additionally, we introduced prompt type augmentation and score perturbation methods for element matching, enhancing the model’s generalization and precision at the element level. Our comprehensive tests on the EvalMuse-40K dataset and the NTIRE 2025 challenge show that iMatch outperforms existing methods in overall and detailed image-text matching, particularly excelling in semantic understanding and fine-grained evaluations. Future work will focus on refining image-text matching strategies and integrating more advanced multimodal models to broaden the applicability of these methods in practical settings.

\clearpage

{
    \small
    \bibliographystyle{ieeenat_fullname}
    \bibliography{main}
}


\end{document}


%% file: main.bbl
\begin{thebibliography}{45}
\providecommand{\natexlab}[1]{#1}
\providecommand{\url}[1]{\texttt{#1}}
\expandafter\ifx\csname urlstyle\endcsname\relax
  \providecommand{\doi}[1]{doi: #1}\else
  \providecommand{\doi}{doi: \begingroup \urlstyle{rm}\Url}\fi

\bibitem[Achiam et~al.(2023)Achiam, Adler, Agarwal, Ahmad, Akkaya, Aleman, Almeida, Altenschmidt, Altman, Anadkat, et~al.]{gpt4}
Josh Achiam, Steven Adler, Sandhini Agarwal, Lama Ahmad, Ilge Akkaya, Florencia~Leoni Aleman, Diogo Almeida, Janko Altenschmidt, Sam Altman, Shyamal Anadkat, et~al.
\newblock Gpt-4 technical report.
\newblock \emph{arXiv preprint arXiv:2303.08774}, 2023.

\bibitem[Arkhipkin et~al.(2023)Arkhipkin, Filatov, Vasilev, Maltseva, Azizov, Pavlov, Agafonova, Kuznetsov, and Dimitrov]{kandinsky3.0}
Vladimir Arkhipkin, Andrei Filatov, Viacheslav Vasilev, Anastasia Maltseva, Said Azizov, Igor Pavlov, Julia Agafonova, Andrey Kuznetsov, and Denis Dimitrov.
\newblock Kandinsky 3.0 technical report.
\newblock \emph{arXiv preprint arXiv:2312.03511}, 2023.

\bibitem[Bai et~al.(2023)Bai, Bai, Yang, Wang, Tan, Wang, Lin, Zhou, and Zhou]{Qwen-VL}
Jinze Bai, Shuai Bai, Shusheng Yang, Shijie Wang, Sinan Tan, Peng Wang, Junyang Lin, Chang Zhou, and Jingren Zhou.
\newblock Qwen-vl: A versatile vision-language model for understanding, localization, text reading, and beyond.
\newblock \emph{arXiv preprint arXiv:2308.12966}, 2023.

\bibitem[Bai et~al.(2025)Bai, Chen, Liu, Wang, Ge, Song, Dang, Wang, Wang, Tang, et~al.]{qwen2.5vl}
Shuai Bai, Keqin Chen, Xuejing Liu, Jialin Wang, Wenbin Ge, Sibo Song, Kai Dang, Peng Wang, Shijie Wang, Jun Tang, et~al.
\newblock Qwen2. 5-vl technical report.
\newblock \emph{arXiv preprint arXiv:2502.13923}, 2025.

\bibitem[blackforestlabs(2024)]{flux}
blackforestlabs.
\newblock Flux1.1.
\newblock \url{https://blackforestlabs.ai/}, 2024.

\bibitem[Chen et~al.(2024{\natexlab{a}})Chen, Wang, Cao, Liu, Gao, Cui, Zhu, Ye, Tian, Liu, et~al.]{internvl2.5}
Zhe Chen, Weiyun Wang, Yue Cao, Yangzhou Liu, Zhangwei Gao, Erfei Cui, Jinguo Zhu, Shenglong Ye, Hao Tian, Zhaoyang Liu, et~al.
\newblock Expanding performance boundaries of open-source multimodal models with model, data, and test-time scaling.
\newblock \emph{arXiv preprint arXiv:2412.05271}, 2024{\natexlab{a}}.

\bibitem[Chen et~al.(2024{\natexlab{b}})Chen, Wu, Wang, Su, Chen, Xing, Zhong, Zhang, Zhu, Lu, et~al.]{internvl}
Zhe Chen, Jiannan Wu, Wenhai Wang, Weijie Su, Guo Chen, Sen Xing, Muyan Zhong, Qinglong Zhang, Xizhou Zhu, Lewei Lu, et~al.
\newblock Internvl: Scaling up vision foundation models and aligning for generic visual-linguistic tasks.
\newblock In \emph{CVPR}, 2024{\natexlab{b}}.

\bibitem[Dosovitskiy et~al.(2020)Dosovitskiy, Beyer, Kolesnikov, Weissenborn, Zhai, Unterthiner, Dehghani, Minderer, Heigold, Gelly, et~al.]{vit}
Alexey Dosovitskiy, Lucas Beyer, Alexander Kolesnikov, Dirk Weissenborn, Xiaohua Zhai, Thomas Unterthiner, Mostafa Dehghani, Matthias Minderer, Georg Heigold, Sylvain Gelly, et~al.
\newblock An image is worth 16x16 words: Transformers for image recognition at scale.
\newblock \emph{arXiv preprint arXiv:2010.11929}, 2020.

\bibitem[DreaminaAI(2023)]{dreamina}
DreaminaAI.
\newblock Dreamina.
\newblock \url{https://dreamina.capcut.com/}, 2023.

\bibitem[Esser et~al.(2024)Esser, Kulal, Blattmann, Entezari, M{\"u}ller, Saini, Levi, Lorenz, Sauer, Boesel, et~al.]{sd3}
Patrick Esser, Sumith Kulal, Andreas Blattmann, Rahim Entezari, Jonas M{\"u}ller, Harry Saini, Yam Levi, Dominik Lorenz, Axel Sauer, Frederic Boesel, et~al.
\newblock Scaling rectified flow transformers for high-resolution image synthesis.
\newblock In \emph{ICML}, 2024.

\bibitem[Feng et~al.(2023)Feng, Gong, Chen, Shen, Liu, and Zhou]{ranni}
Yutong Feng, Biao Gong, Di Chen, Yujun Shen, Yu Liu, and Jingren Zhou.
\newblock Ranni: Taming text-to-image diffusion for accurate instruction following.
\newblock \emph{arXiv preprint arXiv:2311.17002}, 2023.

\bibitem[Han et~al.(2024)Han, Fan, Fu, Li, Li, Cui, Wang, Tai, Sun, Guo, et~al.]{evalmuse}
Shuhao Han, Haotian Fan, Jiachen Fu, Liang Li, Tao Li, Junhui Cui, Yunqiu Wang, Yang Tai, Jingwei Sun, Chunle Guo, et~al.
\newblock Evalmuse-40k: A reliable and fine-grained benchmark with comprehensive human annotations for text-to-image generation model evaluation.
\newblock \emph{arXiv preprint arXiv:2412.18150}, 2024.

\bibitem[Han et~al.(2025)Han, Fan, Kong, Liao, Guo, Li, Timofte, et~al.]{ntire2025text}
Shuhao Han, Haotian Fan, Fangyuan Kong, Wenjie Liao, Chunle Guo, Chongyi Li, Radu Timofte, et~al.
\newblock {NTIRE} 2025 challenge on text to image generation model quality assessment.
\newblock In \emph{Proceedings of the IEEE/CVF Conference on Computer Vision and Pattern Recognition (CVPR) Workshops}, 2025.

\bibitem[He et~al.(2024)He, Fu, Liu, Wang, Xiao, Shu, Wang, Zhang, Yu, Li, et~al.]{mars}
Wanggui He, Siming Fu, Mushui Liu, Xierui Wang, Wenyi Xiao, Fangxun Shu, Yi Wang, Lei Zhang, Zhelun Yu, Haoyuan Li, et~al.
\newblock Mars: Mixture of auto-regressive models for fine-grained text-to-image synthesis.
\newblock \emph{arXiv preprint arXiv:2407.07614}, 2024.

\bibitem[Hessel et~al.(2021)Hessel, Holtzman, Forbes, Bras, and Choi]{clipscore}
Jack Hessel, Ari Holtzman, Maxwell Forbes, Ronan~Le Bras, and Yejin Choi.
\newblock Clipscore: A reference-free evaluation metric for image captioning.
\newblock \emph{arXiv preprint arXiv:2104.08718}, 2021.

\bibitem[Holz(2023)]{midjourney}
David Holz.
\newblock Midjourney.
\newblock \url{https://www.midjourney. com}, 2023.

\bibitem[Hu et~al.(2022)Hu, Shen, Wallis, Allen-Zhu, Li, Wang, Wang, Chen, et~al.]{lora}
Edward~J Hu, Yelong Shen, Phillip Wallis, Zeyuan Allen-Zhu, Yuanzhi Li, Shean Wang, Lu Wang, Weizhu Chen, et~al.
\newblock Lora: Low-rank adaptation of large language models.
\newblock In \emph{ICLR}, 2022.

\bibitem[Hu et~al.(2023)Hu, Liu, Kasai, Wang, Ostendorf, Krishna, and Smith]{tifa}
Yushi Hu, Benlin Liu, Jungo Kasai, Yizhong Wang, Mari Ostendorf, Ranjay Krishna, and Noah~A Smith.
\newblock Tifa: Accurate and interpretable text-to-image faithfulness evaluation with question answering.
\newblock In \emph{ICCV}, 2023.

\bibitem[Kirstain et~al.(2023)Kirstain, Polyak, Singer, Matiana, Penna, and Levy]{pickscore}
Yuval Kirstain, Adam Polyak, Uriel Singer, Shahbuland Matiana, Joe Penna, and Omer Levy.
\newblock Pick-a-pic: An open dataset of user preferences for text-to-image generation.
\newblock In \emph{NeurIPS}, 2023.

\bibitem[Li et~al.(2024{\natexlab{a}})Li, Lin, Pathak, Li, Fei, Wu, Xia, Zhang, Neubig, and Ramanan]{vqascore}
Baiqi Li, Zhiqiu Lin, Deepak Pathak, Jiayao Li, Yixin Fei, Kewen Wu, Xide Xia, Pengchuan Zhang, Graham Neubig, and Deva Ramanan.
\newblock Evaluating and improving compositional text-to-visual generation.
\newblock In \emph{CVPR}, 2024{\natexlab{a}}.

\bibitem[Li et~al.(2024{\natexlab{b}})Li, Kamko, Akhgari, Sabet, Xu, and Doshi]{playgroundv2.5}
Daiqing Li, Aleks Kamko, Ehsan Akhgari, Ali Sabet, Linmiao Xu, and Suhail Doshi.
\newblock Playground v2. 5: Three insights towards enhancing aesthetic quality in text-to-image generation.
\newblock \emph{arXiv preprint arXiv:2402.17245}, 2024{\natexlab{b}}.

\bibitem[Li et~al.(2023)Li, Li, Savarese, and Hoi]{blip2}
Junnan Li, Dongxu Li, Silvio Savarese, and Steven Hoi.
\newblock Blip-2: Bootstrapping language-image pre-training with frozen image encoders and large language models.
\newblock In \emph{ICML}, 2023.

\bibitem[Li et~al.(2024{\natexlab{c}})Li, Zhang, Lin, Xiong, Long, Deng, Zhang, Liu, Huang, Xiao, et~al.]{hunyuandit}
Zhimin Li, Jianwei Zhang, Qin Lin, Jiangfeng Xiong, Yanxin Long, Xinchi Deng, Yingfang Zhang, Xingchao Liu, Minbin Huang, Zedong Xiao, et~al.
\newblock Hunyuan-dit: A powerful multi-resolution diffusion transformer with fine-grained chinese understanding.
\newblock \emph{arXiv preprint arXiv:2405.08748}, 2024{\natexlab{c}}.

\bibitem[Liu et~al.(2023)Liu, Li, Wu, and Lee]{llava}
Haotian Liu, Chunyuan Li, Qingyang Wu, and Yong~Jae Lee.
\newblock Visual instruction tuning.
\newblock In \emph{NeurIPS}, 2023.

\bibitem[Liu et~al.(2024{\natexlab{a}})Liu, Li, Li, and Lee]{llava1.5}
Haotian Liu, Chunyuan Li, Yuheng Li, and Yong~Jae Lee.
\newblock Improved baselines with visual instruction tuning.
\newblock In \emph{CVPR}, 2024{\natexlab{a}}.

\bibitem[Liu et~al.(2024{\natexlab{b}})Liu, Li, Li, Li, Zhang, Shen, and Lee]{llavanext}
Haotian Liu, Chunyuan Li, Yuheng Li, Bo Li, Yuanhan Zhang, Sheng Shen, and Yong~Jae Lee.
\newblock Llavanext: Improved reasoning, ocr, and world knowledge, 2024{\natexlab{b}}.

\bibitem[Lu et~al.(2024{\natexlab{a}})Lu, Liu, Zhang, Wang, Dong, Liu, Sun, Ren, Li, Yang, et~al.]{deepseekvl}
Haoyu Lu, Wen Liu, Bo Zhang, Bingxuan Wang, Kai Dong, Bo Liu, Jingxiang Sun, Tongzheng Ren, Zhuoshu Li, Hao Yang, et~al.
\newblock Deepseek-vl: towards real-world vision-language understanding.
\newblock \emph{arXiv preprint arXiv:2403.05525}, 2024{\natexlab{a}}.

\bibitem[Lu et~al.(2024{\natexlab{b}})Lu, Li, Chen, Xu, Luo, Zhang, and Ye]{ovis2}
Shiyin Lu, Yang Li, Qing-Guo Chen, Zhao Xu, Weihua Luo, Kaifu Zhang, and Han-Jia Ye.
\newblock Ovis: Structural embedding alignment for multimodal large language model.
\newblock \emph{arXiv preprint arXiv:2405.20797}, 2024{\natexlab{b}}.

\bibitem[Radford et~al.(2021)Radford, Kim, Hallacy, Ramesh, Goh, Agarwal, Sastry, Askell, Mishkin, Clark, et~al.]{clip}
Alec Radford, Jong~Wook Kim, Chris Hallacy, Aditya Ramesh, Gabriel Goh, Sandhini Agarwal, Girish Sastry, Amanda Askell, Pamela Mishkin, Jack Clark, et~al.
\newblock Learning transferable visual models from natural language supervision.
\newblock In \emph{ICML}, 2021.

\bibitem[Ramesh et~al.(2022)Ramesh, Dhariwal, Nichol, Chu, and Chen]{dalle}
Aditya Ramesh, Prafulla Dhariwal, Alex Nichol, Casey Chu, and Mark Chen.
\newblock Hierarchical text-conditional image generation with clip latents.
\newblock \emph{arXiv preprint arXiv:2204.06125}, 2022.

\bibitem[Rombach et~al.(2022)Rombach, Blattmann, Lorenz, Esser, and Ommer]{sd1.5}
Robin Rombach, Andreas Blattmann, Dominik Lorenz, Patrick Esser, and Bj{\"o}rn Ommer.
\newblock High-resolution image synthesis with latent diffusion models.
\newblock In \emph{CVPR}, 2022.

\bibitem[Sauer et~al.(2024)Sauer, Lorenz, Blattmann, and Rombach]{sdxl_turbo}
Axel Sauer, Dominik Lorenz, Andreas Blattmann, and Robin Rombach.
\newblock Adversarial diffusion distillation.
\newblock In \emph{ECCV}, 2024.

\bibitem[Shen et~al.(2023)Shen, Song, Tan, Li, Lu, and Zhuang]{hugginggpt}
Yongliang Shen, Kaitao Song, Xu Tan, Dongsheng Li, Weiming Lu, and Yueting Zhuang.
\newblock Hugginggpt: Solving ai tasks with chatgpt and its friends in huggingface.
\newblock In \emph{NeurIPS}, 2023.

\bibitem[Team et~al.(2023)Team, Anil, Borgeaud, Alayrac, Yu, Soricut, Schalkwyk, Dai, Hauth, Millican, et~al.]{gemini}
Gemini Team, Rohan Anil, Sebastian Borgeaud, Jean-Baptiste Alayrac, Jiahui Yu, Radu Soricut, Johan Schalkwyk, Andrew~M Dai, Anja Hauth, Katie Millican, et~al.
\newblock Gemini: a family of highly capable multimodal models.
\newblock \emph{arXiv preprint arXiv:2312.11805}, 2023.

\bibitem[Team(2024)]{kolors}
Kolors Team.
\newblock Kolors: Effective training of diffusion model for photorealistic text-to-image synthesis.
\newblock \emph{arXiv preprint}, 2024.

\bibitem[Wei et~al.(2022)Wei, Wang, Schuurmans, Bosma, Xia, Chi, Le, Zhou, et~al.]{cot}
Jason Wei, Xuezhi Wang, Dale Schuurmans, Maarten Bosma, Fei Xia, Ed Chi, Quoc~V Le, Denny Zhou, et~al.
\newblock Chain-of-thought prompting elicits reasoning in large language models.
\newblock In \emph{NeurIPS}, 2022.

\bibitem[Wiles et~al.(2024)Wiles, Zhang, Albuquerque, Kaji{\'c}, Wang, Bugliarello, Onoe, Knutsen, Rashtchian, Pont-Tuset, et~al.]{gecko}
Olivia Wiles, Chuhan Zhang, Isabela Albuquerque, Ivana Kaji{\'c}, Su Wang, Emanuele Bugliarello, Yasumasa Onoe, Chris Knutsen, Cyrus Rashtchian, Jordi Pont-Tuset, et~al.
\newblock Revisiting text-to-image evaluation with gecko: On metrics, prompts, and human ratings.
\newblock \emph{arXiv preprint arXiv:2404.16820}, 2024.

\bibitem[Wu et~al.(2023{\natexlab{a}})Wu, Yin, Qi, Wang, Tang, and Duan]{visualchatgpt}
Chenfei Wu, Shengming Yin, Weizhen Qi, Xiaodong Wang, Zecheng Tang, and Nan Duan.
\newblock Visual chatgpt: Talking, drawing and editing with visual foundation models.
\newblock \emph{arXiv preprint arXiv:2303.04671}, 2023{\natexlab{a}}.

\bibitem[Wu et~al.(2024)Wu, Zhang, Zhang, Chen, Liao, Li, Gao, Wang, Zhang, Sun, et~al.]{qalign}
Haoning Wu, Zicheng Zhang, Weixia Zhang, Chaofeng Chen, Liang Liao, Chunyi Li, Yixuan Gao, Annan Wang, Erli Zhang, Wenxiu Sun, et~al.
\newblock Q-align: Teaching lmms for visual scoring via discrete text-defined levels.
\newblock In \emph{ICML}, 2024.

\bibitem[Wu et~al.(2023{\natexlab{b}})Wu, Hao, Sun, Chen, Zhu, Zhao, and Li]{hpsv2}
Xiaoshi Wu, Yiming Hao, Keqiang Sun, Yixiong Chen, Feng Zhu, Rui Zhao, and Hongsheng Li.
\newblock Human preference score v2: A solid benchmark for evaluating human preferences of text-to-image synthesis.
\newblock \emph{arXiv preprint arXiv:2306.09341}, 2023{\natexlab{b}}.

\bibitem[Xu et~al.(2023)Xu, Liu, Wu, Tong, Li, Ding, Tang, and Dong]{imagereward}
Jiazheng Xu, Xiao Liu, Yuchen Wu, Yuxuan Tong, Qinkai Li, Ming Ding, Jie Tang, and Yuxiao Dong.
\newblock Imagereward: Learning and evaluating human preferences for text-to-image generation.
\newblock In \emph{NeurIPS}, 2023.

\bibitem[Yarom et~al.(2023)Yarom, Bitton, Changpinyo, Aharoni, Herzig, Lang, Ofek, and Szpektor]{vq2}
Michal Yarom, Yonatan Bitton, Soravit Changpinyo, Roee Aharoni, Jonathan Herzig, Oran Lang, Eran Ofek, and Idan Szpektor.
\newblock What you see is what you read? improving text-image alignment evaluation.
\newblock In \emph{NeurIPS}, 2023.

\bibitem[Ye et~al.(2024)Ye, Xu, Liu, Hu, Yan, Qian, Zhang, Huang, and Zhou]{mplugowl3}
Jiabo Ye, Haiyang Xu, Haowei Liu, Anwen Hu, Ming Yan, Qi Qian, Ji Zhang, Fei Huang, and Jingren Zhou.
\newblock mplug-owl3: Towards long image-sequence understanding in multi-modal large language models.
\newblock In \emph{ICLR}, 2024.

\bibitem[Zhao et~al.(2024)Zhao, Huang, Hu, Wang, Mao, Zhang, Jiang, Wu, Ai, Wang, Zhou, and Chen]{ms-swift}
Yuze Zhao, Jintao Huang, Jinghan Hu, Xingjun Wang, Yunlin Mao, Daoze Zhang, Zeyinzi Jiang, Zhikai Wu, Baole Ai, Ang Wang, Wenmeng Zhou, and Yingda Chen.
\newblock Swift:a scalable lightweight infrastructure for fine-tuning, 2024.

\bibitem[Zhu et~al.(2023)Zhu, Chen, Shen, Li, and Elhoseiny]{minigpt-4}
Deyao Zhu, Jun Chen, Xiaoqian Shen, Xiang Li, and Mohamed Elhoseiny.
\newblock Minigpt-4: Enhancing vision-language understanding with advanced large language models.
\newblock \emph{arXiv preprint arXiv:2304.10592}, 2023.

\end{thebibliography}
